# Toward Reliable Tea Leaf Disease Diagnosis Using Deep Learning Model: Enhancing Robustness With Explainable AI and Adversarial Training

Samanta Ghosh
Dept. of CSE
East West University
Dhaka, Bangladesh
Samantaewu28@gmail.com

Jannatul Adan Mahi
Dept. of CSE
East West University
Dhaka, Bangladesh
jannatuladanmahi@gmail.com

Shayan Abrar
Dept. of CS
American International University- Bangladesh (AIUB)
Dhaka, Bangladesh
shayanabrar7@gmail.com

Md Parvez Mia
Dept. of CSE
East West University
Dhaka, Bangladesh
mdparvezmia999@gmail.com

Asaduzzaman Rayhan
Dept. of CSE
Ahsanullah University Of Science And Technology
Dhaka, Bangladesh
asaduzzamanrayhan15@gmail.com

Abdul Awal Yasir
Dept. of CSE
North South University
Dhaka, Bangladesh
Yasir.nafis201@gmail.com

Asaduzzaman Hridoy
Dept. of ICT
Bangladesh University of Professional
Dhaka, Bangladesh
asaduzzamanhridoyice@gmail.com

*Abstract*— Tea is a valuable asset for the economy of Bangladesh. So, tea cultivation plays an important role to boost the economy. These valuable plants are vulnerable to various kinds of leaf infections which may cause less production and low quality. It is not so easy to detect these diseases manually. It may take time and there could be some errors in the detection. Therefore, the purpose of the study is to develop an automated deep learning model for tea leaf disease classification based on the teaLeafBD dataset so that anyone can detect the diseases more easily and efficiently. There are 5,278 high-resolution images in this dataset. The images are classified into seven categories. Six of them represents various diseases and the rest one represents healthy leaves. The proposed pipeline contains data preprocessing, data splitting, adversarial training, augmentation, model training, evaluation, and comprehension made possible with Explainable AI strategies. DenseNet201 and EfficientNetB3 were employed to perform the classification task. To prepare the model more robustly, we applied adversarial training so it can operate effectively even with noisy or disturbed inputs. In addition, Grad-CAM visualization was executed to analyze the model's predictions by identifying the most influential regions of each image. Our experimental outcomes revealed that EfficientNetB3 achieved the highest classification accuracy of 93%, while DenseNet201 reached 91%. The outcomes prove that the effectiveness of the proposed approach can accurately detect tea leaf diseases and provide a practical solution for advanced agricultural management.

## I. Introduction

Tea is one of the most significant cash crops in Bangladesh, contributing primarily to the national economy and employing a large portion of the rural population. The country ranks among the top tea producers internationally, with the Sylhet and Chittagong regions being the primary cultivation zones. However, tea plantations in Bangladesh often suffer from various leaf diseases such as blister blight, red dust, and gray blight, which severely affect both the yield and quality of the crop. Traditional methods of disease identification, which rely heavily on visual inspection by agricultural experts, are often subjective, time-consuming, and impractical for widespread implementation across large tea plantations. These systems can automate the identification process, reduce human error, and deliver faster decision-making support to growers.

Recent advancements in neural networks have significantly improved image-based disease classification by enabling automated feature extraction and high-level pattern recognition. The study by Saikat Datta et al. suggested a novel deep CNN model for detecting and classifying tea leaf disease. They used a dataset of 5,867 images, which covers five disease types, and they are Algal Spot, Brown Blight, Gray Blight, Helopeltis, and Red Spot, along with healthy leaves. They proposed a 16-layer CNN model that achieved higher accuracy, outperforming SVM and ANN [1].

A comprehensive review of convolutional neural network applications in plant leaf disease classification was presented, which discusses CNN principles and the advantages of deep learning over traditional machine learning approaches. This review emphasizes that CNN-based systems offer state-of-the-art accuracy for plant disease detection, whereas their effectiveness in real-world scenarios depends heavily on dataset diversity and model generalization [2].

Jing Chen et al. developed LeafNet, a modified AlexNet CNN for tea leaf disease recognition. Performance was compared against SVM and MLP using Dense SIFT features, and LeafNet outperformed SVM and MLP [3]. The study concludes that CNN-based architectures like LeafNet are more efficient for complex visual disease classification tasks in tea cultivation.

This research adopts two advanced deep learning architectures, DenseNet201 and EfficientNetB3, to classify tea leaf diseases. These models are known for their higher accuracy, efficient feature extraction, and strong performance in image-based classification tasks. Such characteristics make them well-suited for detecting and classifying various tea leaf diseases from agricultural images. This study not only explores performance in terms of classification accuracy but also emphasizes robustness, scalability, and field applicability.

Research Questions:
- Can data preprocessing and augmentation strategies improve the generalization performance of deep learning models in tea leaf disease detection?
- To what extent can explainable AI enhance the interpretability of deep learning models in tea leaf disease detection?

## II. RELATED WORKS

The authors [4] fine-tuned transfer-learning models DenseNet201, InceptionV3, and Xception on a real-world dataset of 1006 coffee leaf images across 10 nutrient-deficiency classes. The Xception model yielded the highest accuracy at 83.32 %, not over 90 %. Grad CAM visualization was employed to highlight leaf regions influencing model decisions, enhancing interpretability. While the approach is promising, the accuracy remains modest, limiting its readiness for agricultural deployment without further robustness testing and field validation.

Chen et al. [3] proposed a CNN-based method called LeafNet for recognizing seven types of tea leaf diseases. They used a self-built image dataset containing diseased tea leaves like white spot, red leaf spot, and gray blight. LeafNet was trained and tested on this dataset and achieved a high classification accuracy of 90.16%, significantly outperforming traditional classifiers such as SVM (60.62%) and MLP (70.77%). The model also showed strong class-wise sensitivity, with up to 98.32% for the bird's-eye spot. A confusion matrix was used to evaluate per-class performance. The authors highlighted that CNN models like LeafNet are more effective than handcrafted feature methods in visual plant disease recognition [5]. The study highlights that deep learning can be a powerful tool for real-time disease detection in tea leaves, contributing to smart agriculture and early disease management.

A study focused on detecting and classifying tea leaf diseases using deep learning [6]. A Convolutional Neural Network (CNN) was applied to tea leaf images collected under various lighting and environmental conditions. The CNN was compared with traditional models such as Support Vector Machine (SVM) and k-Nearest Neighbors (k-NN). Results indicate that CNN achieved the highest accuracy of 92.59%, outperforming the other methods. The research demonstrated CNN's ability to extract features effectively for disease detection, though challenges remained with overlapping symptoms and limited dataset size.

A tea leaf disease recognition system was developed using Neural Network Ensemble (NNE) with Negative Correlation Learning (NCL) to classify infected leaves [7]. The process includes image preprocessing, thresholding, normalization, feature extraction, and classification. The model was trained on 50 images from five disease categories, achieving a maximum recognition accuracy of 91%. Compared to the Scale Conjugate Gradient (SCG), the NCL-based system demonstrated faster learning and higher accuracy. The input images were resized to 30×33 pixels, with the final model using ten neural networks and feature extraction for optimal performance. The system offers a low-cost, non-invasive solution for early detection of tea leaf diseases and has potential for a real-time mobile application in agricultural monitoring.

This study compares multiple machine learning algorithms for detecting seven types of tea leaf diseases, including anthracnose, white spot, bird's-eye, red leaf spot, brown blight, grey blight, and algal leaf [8]. Using the Tea Disease Dataset of 1,106 images, models such as Extra Tree Classifier (ETC), Support Vector Classifier (SVC), Random Forest (RF), XGBoost (XGB), Decision Tree (DT), and Convolutional Neural Network (CNN) were evaluated. Performance metrics included accuracy, precision, recall, and F1-score. Results showed ETC achieved the highest accuracy of 77.47%, while CNN performed the worst at 59.08%. The findings highlight ETC as the most effective method for tea leaf disease detection.

Tea leaf disease detection has been studied using various CNN architectures. AlexNet achieved 83.15% accuracy, while ResNet101 reached 82.58% and GoogLeNet 78.09%. Lightweight models such as ShuffleNet 84.27% and Darknet53 82.02% performed well for faster detection, whereas MobileNetV2 obtained 76.97%. EfficientNetB0 showed lower performance, 66.29%, indicating dataset-specific sensitivity [9] [10]. Hybrid approaches in related works often fuse features from multiple CNNs with classifiers like SVM or LDA, leading to improved accuracy. In this study's context, a proposed fusion of EfficientNetB0, ResNet101, and ShuffleNet achieved the highest accuracy of 91.3%, outperforming all individual models and showing the benefit of feature fusion.

## III. METHODOLOGY

Fig. 1 illustrates the entire process of the proposed system for tea leaf disease classification using deep learning.

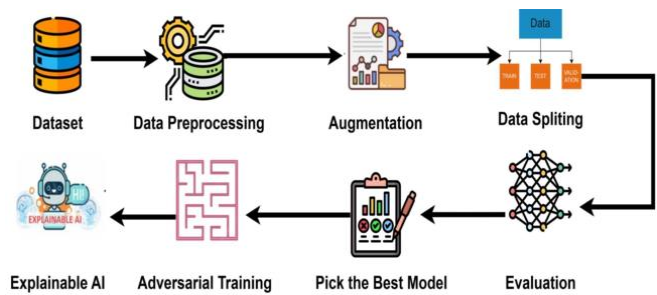

Fig. 1. Methodology

Dataset acquisition is the beginning of the process, where raw images of tea leaves are collected. Data preprocessing is applied to these images afterward, involving steps like normalization, resizing, and noise removal to organize the data for modeling. Augmentation approaches [11] like flipping, rotation, and scaling are applied to artificially enhance the dataset and improve model generalization during the augmentation phase. The updated dataset is then split into training (70%), validation (20%), and testing (10%) subsets. Different deep learning models are trained and evaluate the performance using the standard metrics. EfficientNetB3 and DenseNet201 are selected because they obtained the best proficiency on our datasets. Adversarial training is used to enhance the stability of a model. This ensures stable performance under noisy data. Lastly GRAD-CAM is used to highlight the important region of an image and explain the model's decision. This workflow ensures the development of a robust, accurate and interpretable framework for tea leaf disease detection.

In this study, the data dataset shown in Fig. 2 is collected from Mendeley data [12]. The dataset contains 5,278 images. Each image measuring 1200x1600 pixels with RGB color mode. Some preprocessing techniques are applied to improve the image quality. Images were resized to 224x224 pixels. The dataset contains seven classes. These are Tea algal leaf spot (418), Brown Blight (508), Gray Blight (1013), Helopeltis (607), Red spider (515), Green mirid bug (1282), Healthy Leaf (935) [10]. Tea algal leaf spot is a fungal infection. These lead to form small greenish-gray spot and affect the leaf quality. Brown blight is also known as fungal disease leading to leaf drying and reduced tea yield. This creates brown patches on the leaves. Gray blight produces grayish spots on leaves and effects on the plant's growth. Helopeltis disease leads to discolored spots on leaves and affects the quality of the tea. Red spider is an infection for which the quality and production of the tea leaves may reduce. This infection is caused by red spider mites. This insect harms the tea leaves by damaging the sap and leads to bronzing and wintergreen mirid bug disease happens when a pest or insects eats tea leaves, making holes and spots. These leads to reduced tea yield. Healthy leaf illustrates the normal condition of a leaf which is clean, intake and unaffected by any disease or insects.

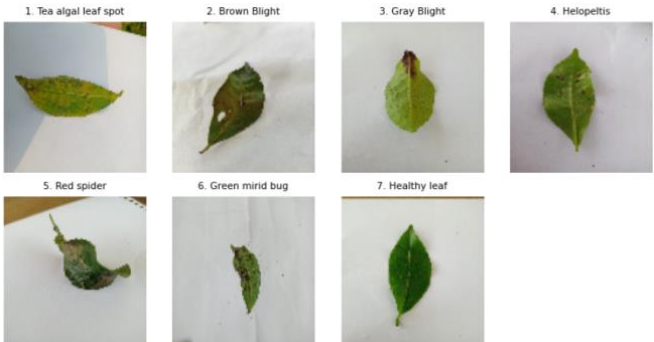
Fig. 2. Dataset Sample

### A. DenseNet201

Densenet201 is a deep learning model that is designed to optimize feature propagation and reduce the number of parameters through smart layer connections [13]. This model has 201 layers. These layers help to learn difficult visual features from high-resolution images. To do this, this model does not even need excessive computing power. Fig. 3 shows the structure of the Densenet201 model. To process an image, this model passes the image through different dense blocks. These blocks are indicated by unique color segments. Every dense block contains multiple convolutional layers. The output of each layer relates to the outputs of all prior layers. The feature map shape is reduced by transition layers between dense blocks using pooling and 1x1 convolutions, enhancing the model's intensity and complexity. For regularization, the extracted features are compressed and fully fed into connected layers with ReLU activation and dropout. Finally, the model predicts the class, making accurate classification of tea leaf images over multiple categories of diseases.

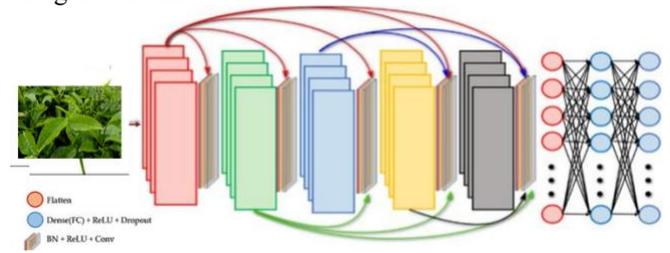
Fig. 3. Proposed DenseNet201 Architecture

### B. EfficientNetB3

EfficientNetB3 [14] is a strong CNN model that uses multi-dimensional scaling to adjust its depth, width, and resolution efficiently. This model is particularly effective for image classification tasks that involve fine-grained visual differences. To extract rich features while being parameter-efficient, it uses MBConv layers and Swish activation, and this makes the model optimal for real-time, resource-limited applications like agricultural disease detection on mobile devices.

Fig. 4 shows the architecture of EfficientNetB3. It initiates with an input image of size 224×224×3, widely adopted in image classification tasks. The model first applies a regular 3×3 convolution layer, then uses various MBConv blocks that have distinct kernel sizes (3×3 or 5×5), expansion factor, and resolutions. These blocks perform depth wise split convolutions, and skip connections are incorporated to keep the flow of information intact. As the image moves through each block, it reduces in size while the details it learns more deeply, helping the model extract useful features efficiently. The final layers include a 1×1 convolution, then global average pooling, and at the end, a fully connected layer is used that produces the class probabilities. This modular design helps EfficientNetB3 learn features gradually, making it powerful for detailed tasks like detecting specific tea leaf diseases.

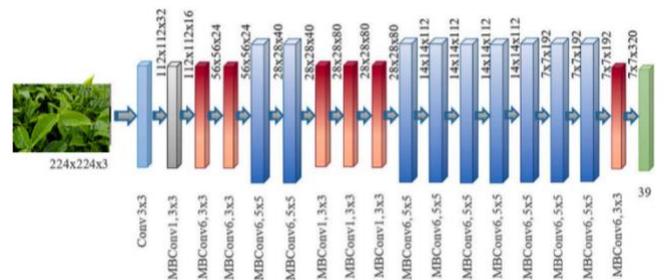
Fig. 4. Proposed EfficientNetB3 Architecture

## IV. EXPERIMENTAL ANALYSIS

### A. DenseNet201

The curves in the Fig. 5 shows that training accuracy is being increased from 70% to nearly 100% by the end of the training. On the other hand, validation accuracy starts from 52% and then it starts to fall below 20% in the initial epochs. Eventually, it stabilizes around 91%. This suggests that there is an instability in the validation stage during early training. The loss curve shows the training loss is being decreased gradually which indicates the learning progress. Validation loss increases at epoch 4.

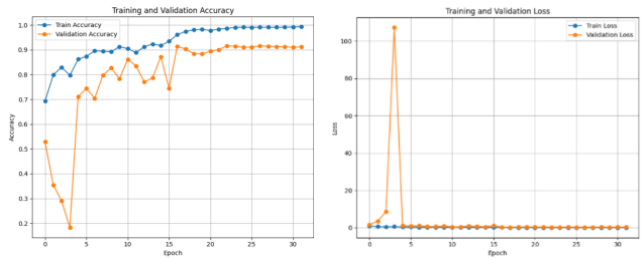
Fig. 5. Loss and accuracy curve of DenseNet201 architecture

It even exceeded 100 before rapidly decreasing and eventually stabilizes near zero. This sudden increase indicates that there is a possible data irregularity during that epoch. Except this, the tiny gap between training and validation metrics tells that the model eventually learned generalization well after recovering from the initial instability

As shown in Fig. 6, the confusion matrix of the DenseNet201 model illustrates that it accurately classifies all seven types of tea leaves. A large number of samples were labeled correctly, having the most accurate results for Green mirid bug (116) and Gray Blight (96). A few misclassifications were observed between Brown Blight and Tea algal leaf spot (8 instances), and a few Helopeltis samples were wrongly identified as red spider or Green mirid bug. Robust leaves were predominantly recognized, with 91 properly categorized instances. Overall, the matrix highlights the model's remarkable ability to distinguish both diseased and healthy leaves, showing DenseNet201 is well-suited for automatic tea leaf disease detection.

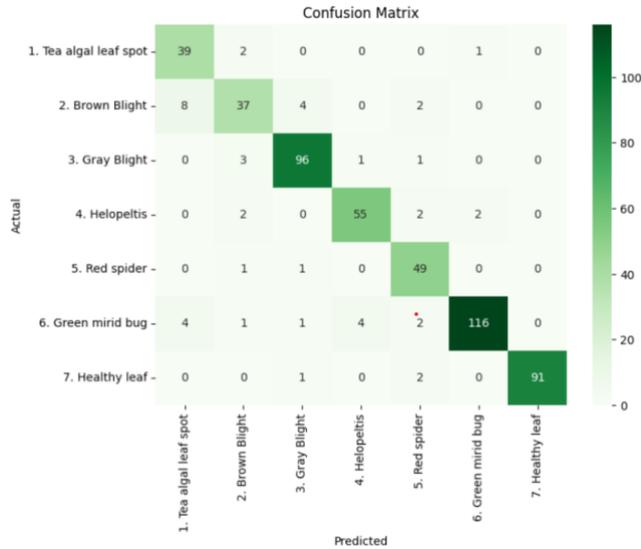
Fig. 6. Confusion Matrix of DenseNet201

Table 1 shows the hyperparameter settings used for training the DenseNet201 model.

TABLE I HYPERPARAMETER TUNING DENSENET201

| **Batch size** | 32 | **Loss function** | categorical_crossentropy |
|---|---|---|---|
| **Learning rate** | 0.0001 | **Number of epochs** | 32 |
| **Optimizer** | Adam | **Patience** | 15 |

The network was trained with a batch size of 32, using the categorical cross-entropy loss function and the Adam optimizer. The learning rate was set to 0.0001 to ensure gradual and smooth convergence. The model was trained for 32 epochs, using early stopping applied with a patience value of 15 to prevent overfitting and preserve the best weights during training.

*B. EfficientNetB3*

The accuracy curve shows in the Fig. 7 that the performance of the model is getting better rapidly during the initial epochs for training and validation datasets. Training accuracy starts at 45% and rises quickly. It reached 90% on the fifth epoch. Eventually, it achieved 99% accuracy. Validation accuracy also improves sharply at the beginning, stabilizing around 93% after the initial epochs. The loss curve helps this by showing a steep decline in training and validation loss at the initial stage of the training. Training loss decreases up to 1.75 to nearly zero. In the meantime, validation loss decreases from 0.75 to nearly 0.26 and stays steady. This pattern proves that the model has learned the patterns in the training data very effectively. Therefore, this model is able to maintain strong generalization to unseen data. The tiny gap between training and validation metrics shows a small overfitting but not to a harmful level. It makes the model accurate and reliable for practical implementation.

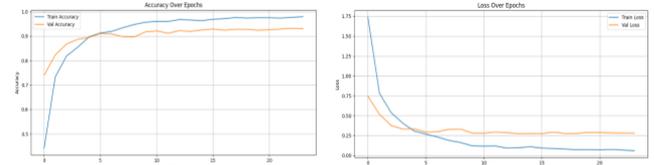
Fig. 7. Training and validation accuracy and loss curve of EfficientNetB3

According to Fig. 8, the confusion matrix explains high classification accuracy over all categories. 125 Green mirid bug, 94 Gray Blight, and 91 Healthy leaf samples were successfully recognized by the model. Five Brown Blight samples were predicted as Tea algal leaf spot, and four Tea algal spot samples were improperly categorized as Brown Blight. Moreover, one Helopeltis sample was assigned to the Red spider class in error. Regardless of these minor issues, a strong ability to differentiate between healthy and diseased leaves was illustrated by EfficientNetB3, proving its effectiveness for automated tea leaf disease recognition.

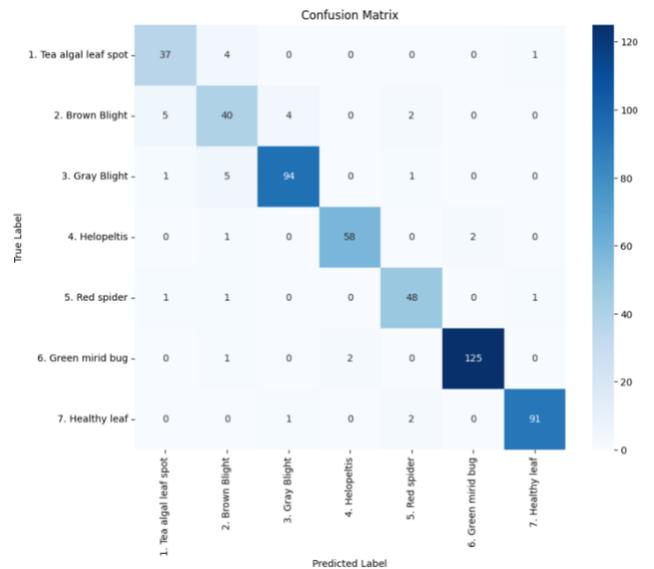
Fig. 8. Confusion Matrix of EfficientNetB3

Table 2 outlines the hyperparameter configuration adopted for the EfficientNetB3 model. The loss function was categorical cross-entropy, the Adam optimizer was employed, and the batch size was set to 32. The learning rate was defined at 0.0001 for managed optimization. The model was trained for 24 epochs, and early stopping was applied using a patience value of 15 when validation stopped improving, saving time and preventing overfitting.

TABLE II HYPERPARAMETER TUNING OF EFFICIENTNETB3

| Batch size | 32 | Loss function | categorical_crossentropy |
|---|---|---|---|
| Learning rate | 0.0001 | Number of epochs | 24 |
| Optimizer | Adam | Patience | 15 |

The Table 3 represents the outcomes of adversarial training among different epsilon (ε) values that define the intensity of adversarial perturbations. This is applied to enhance the stability of the model. EfficientNetB3 were trained with distinctly designed challenging inputs during this process. It helped the model to learn more stable features and deal with challenging inputs. The model reached its maximum validation accuracy of 81.9% with a validation loss of 0.5108 for ε = 0 (no noise) and requiring 88 epochs to converge. Adding small perturbations (ε between 0.1 and 0.2) increased performance and helped the model reach its best accuracy much faster, generally within 24-39 epochs. The highest accuracy was achieved at ε = 0.12, it was 84.83%, and values like 0.14, 0.18, and 0.2 also stayed above 84%. Validation loss stayed around 0.53-0.57 for all ε values, indicating stable generalization. In general, model robustness and accuracy improved, and faster convergence was achieved by adding moderate adversarial noise, with ε= 0.12 proving to be the top-performing setting in this experiment. Therefore, EfficientNetB3 maintained high accuracy even on challenging instances.

TABLE III PERFORMANCE OF EFFICIENTNETB3 FOR ADVERSARIAL EXAMPLES

| Epsilon Value (ε) | Validation Loss | Validation Accuracy | Optimal Epochs |
|---|---|---|---|
| 0 | 0.5108 | 81.9 | 88 |
| 0.1 | 0.5483 | 83.79 | 24 |
| 0.12 | 0.567 | 84.83 | 39 |
| 0.14 | 0.5388 | 84.17 | 24 |
| 0.16 | 0.5596 | 83.89 | 29 |
| 0.18 | 0.5384 | 84.08 | 24 |
| 0.2 | 0.5451 | 84.17 | 24 |

V. EXPLAINABLE AI

A. SHAP

Fig. 9 presents the implementation of Grad-CAM as a component of the explainable AI of this study. Shown on the left is the original input image, which is a tea leaf affected by disease, as shown by distinct visual cues of discoloration and lesions. The right image presents the Grad-CAM heatmap, which highlights the parts of the image that made the most meaningful contribution to the model's prediction [15]. The heatmap applies a color gradient, where red and yellow areas show high relevance, and blue and green indicate lower relevance. Significantly, the model targets the lower left region of the leaf where the disease is clearly noticeable. The alignment between the elements targeted by the model and the actual diseased area confirms that relevant features are being detected by the deep learning model, and decisions are benignly generated based on insightful visual patterns. The use of Grad-CAM not only provides transparency into the model's reasoning mechanism but also enhances interpretability and dependability, which are necessary for real-world use in agriculture and plant disease management.

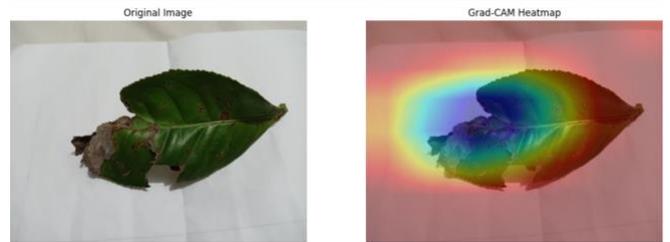

Fig. 9. Tea leaf disease classification using Grad-CAM

VI. COMPARISON

Table 4 shows the training and test accuracy of Densetnet201 and EfficientNetB3. DenseNet201 showed an excellent accuracy, achieving 99.99% training accuracy. It also achieved 91.10% test accuracy. It showed excellent learning, but a small drop in unseen data. On the other hand, EfficientNetB3 achieved 98% training accuracy and 93.43% test accuracy, which indicates better generalization. DenseNet201 fits the training set more closely than EfficientNetB3. Meanwhile, EfficientNetB3 can maintain stronger performance on test data.

TABLE IV ACCURACY COMPARISON BETWEEN THE TWO ARCHITECTURES

| Model name | Training accuracy (%) | Test accuracy (%) |
|---|---|---|
| Densenet201 | 99.99 | 91.10 |
| EfficientNetB3 | 98 | 93.43 |

Table 5 shows the detailed classification outcomes for DenseNet201 and EfficientNetB3 on seven varieties of tea leaf conditions in the TeaLeaf dataset: tea algal leaf spot, brown blight, gray blight, helopeltis, red spider, green mirid bug, and healthy leaf.

TABLE V COMPARISON BETWEEN THE TWO ARCHITECTURES

| Model Name | Class | Precision | Recall | F1-Score | Accuracy |
|---|---|---|---|---|---|
| Dense Net201 | Tea algal leaf spot | 0.88 | 0.86 | 0.87 | 91% |
|  | Brown Blight | 0.82 | 0.80 | 0.81 |  |
|  | Gray Blight | 0.91 | 0.91 | 0.91 |  |
|  | Helopeltis | 0.93 | 0.89 | 0.91 |  |
|  | Red spider | 0.88 | 0.86 | 0.87 |  |
|  | Green mirid bug | 0.91 | 0.97 | 0.94 |  |
|  | Healthy leaf | 0.99 | 0.96 | 0.97 |  |
| EfficientNetB3 | Tea algal leaf spot | 0.84 | 0.88 | 0.86 | 93% |
|  | Brown Blight | 0.77 | 0.78 | 0.78 |  |
|  | Gray Blight | 0.95 | 0.93 | 0.94 |  |
|  | Helopeltis | 0.97 | 0.95 | 0.96 |  |
|  | Red spider | 0.91 | 0.94 | 0.92 |  |
|  | Green mirid bug | 0.98 | 0.98 | 0.98 |  |
|  | Healthy leaf | 0.98 | 0.97 | 0.97 |  |

Including precision, recall, and F1-score for each class, along with the overall accuracy of the models, is reported by the metrics. DenseNet201 achieved an overall accuracy of 91%, doing exceptionally well in detecting healthy leaves (F1-score: 0.97) and green mirid bugs (F1-score: 0.94). EfficientNetB3 achieved a higher overall accuracy of 93% compared to DenseNet201, performing especially well in identifying green mirid bugs (F1-score: 0.98) and helopeltis

(F1-score: 0.96), which highlights its superior classification capability. The overall performance of the EfficientNetB3 model is better than the DenseNet201 model. This highlights that EfficientNetb3 model gives better scalability and consistency results for the given dataset.

Table 6 shows the comparison performance of the proposed models with several existing approaches reported in the literature. Methods such as Xception, LeafNet, Support Vector Machine, Neural Network Ensemble, Extra Tree Classifier (ETC), and ShuffleNet are mentioned with their respective accuracies. In previous work, the Support Vector Machine achieved 92.59%, which was the highest accuracy. Our proposed DenseNet201 model reached 91% accuracy, while the EfficientNetB3 model achieved the highest accuracy overall at 93%, and it did better than most earlier methods, showing that it works well for classifying tea leaf diseases.

TABLE VI COMPARISON WITH PREVIOUS WORKS

| Author | Method | Accuracy (%) |
|---|---|---|
| Ahammed et al. [4] | Xception | 83.32 |
| Chen et al. [3] | LeafNet | 90.16 |
| Chakraborty et al. [6] | Support Vector Machine | 92.59 |
| B. C. Karmokar et al. [7] | Neural Network Ensemble | 91 |
| J. RESTI [8] | Extra Tree Classifier (ETC) | 77.47 |
| N. Yucel and M. Yildirim [9] | ShuffleNet | 91.3 |
| Proposed Model | DenseNet201 | 91 |
| | EfficientNetB3 | 93 |

## VII. CONCLUSION

In order to benefit farmers and tea cultivators in early detection of disease and yield preservation, this research tried to create a deep learning method for precisely and effectively detecting diseases of the tea leaf. Small datasets, insufficient quality of images inadequate disease class coverage were frequent issues in previously tea leaf disease detection research, which diminished model generalization. We used a dataset of 5,278 images from seven diseases and appropriate leaf classes to overcome these constraints, ensuring more effective variation as well as balanced visualization. After a strong evaluation, we determined the DenseNet201 and EfficientNetB3 architectures, identifying that EfficientNetB3 accomplished the best with an accuracy of 91% and 93% respectively. Our technique depicts immense potential for being incorporated into a web-based or mobile application, providing farmers, tea estate managers, and agricultural officers with easily accessible, real-time disease detection. Future research may investigate further improvement for real-time prediction and robustness under different environmental circumstances, even though the outcome of the model is beneficial. All things considered, our study represents an important milestone in the implementation of deep learning for realistic, feasible, and scalable tea leaf disease detection, which improves both the standard and productivity of tea cultivation.